%% file: ijcb_2026.tex
\documentclass[10pt,twocolumn,letterpaper]{article}

\usepackage{cvpr}
\usepackage{times}
\usepackage{epsfig}
\usepackage{graphicx}
\usepackage{amsmath}
\usepackage{amssymb}
\usepackage{stfloats}
\usepackage{booktabs} 
\usepackage[colorlinks=true,
            linkcolor=blue,
            citecolor=blue,
            urlcolor=blue]{hyperref} %
\usepackage[table]{xcolor}            
\usepackage{pifont}                   
\newcommand{\cmark}{\ding{51}}        
\usepackage{bm}
\usepackage{enumitem}

\usepackage{fancyhdr}

\fancypagestyle{acceptedpaper}{
    \fancyhf{}
    \fancyhead[R]{%
        \begin{minipage}{0.72\textwidth}
            \raggedleft
            \footnotesize\itshape
            Accepted at the IEEE/IAPR International Joint Conference
            on Biometrics (IJCB 2026)
        \end{minipage}
    }

}

\begin{document}

\title{Uncertainty-Aware Deepfake Detection via Multi-View Structural Learning}
\author{Muhammad Umar Farooq
\and Kutub Uddin
\and Awais Khan
\and Khalid Malik\\
College of Innovation and Technology\\
University of Michigan, Flint, MI 48502\\
{\tt\small \{mufarooq, kutub, mawais, drmalik\}@umich.edu}
}


\maketitle

\pagestyle{acceptedpaper}
\thispagestyle{acceptedpaper}
\input{secs/abstract}

\input{secs/introduction}

\input{secs/literature_review}
\input{secs/proposed_method}
\input{secs/experiments}

\section*{Acknowledgment}\vspace{-5pt}
This material is based upon work supported by the National Science Foundation (NSF) under Grant number 2409577. Any opinions, findings, and conclusions, or recommendations expressed in this material are those of the author(s) and do not necessarily reflect the views of the NSF.

{\small
\bibliographystyle{unsrt}
\bibliography{references}
}

\end{document}

%% file: secs/abstract.tex
\begin{abstract}

Security-critical biometric and forensic applications require accurate predictions and reliable confidence estimates, particularly under distribution shift. This challenge is especially acute for deepfake detection, where foundation-model-based detectors often exhibit overconfident predictions on out-of-distribution manipulations, which limits their suitability for operational deployment. We propose an uncertainty-aware deepfake detection framework that identifies manipulations through inconsistencies across complementary evidence sources. The framework integrates three streams: a visual stream based on an adapted CLIP encoder, a semantic stream that models consistency among facial attributes through differentiable constraints, and a structural stream that captures class-dependent dependency patterns between semantic and forensic features. To effectively combine these signals, we introduce Inter-Branch Disagreement Calibration (IBDC), a disagreement-aware uncertainty modeling mechanism that links predictive uncertainty to conflicts among evidence streams. Extensive cross-dataset experiments using FaceForensics++ as the training source demonstrate that the proposed framework achieves state-of-the-art generalization across multiple out-of-distribution benchmarks while consistently improving calibration and selective prediction performance. These results show that combining complementary evidence with disagreement-aware uncertainty provides a robust foundation for trustworthy and well-calibrated deepfake detection under distribution shift.
\end{abstract}
\vspace{-20pt}

%% file: secs/introduction.tex
\begin{figure}[t]
  \centering
  \includegraphics[width=1\linewidth]{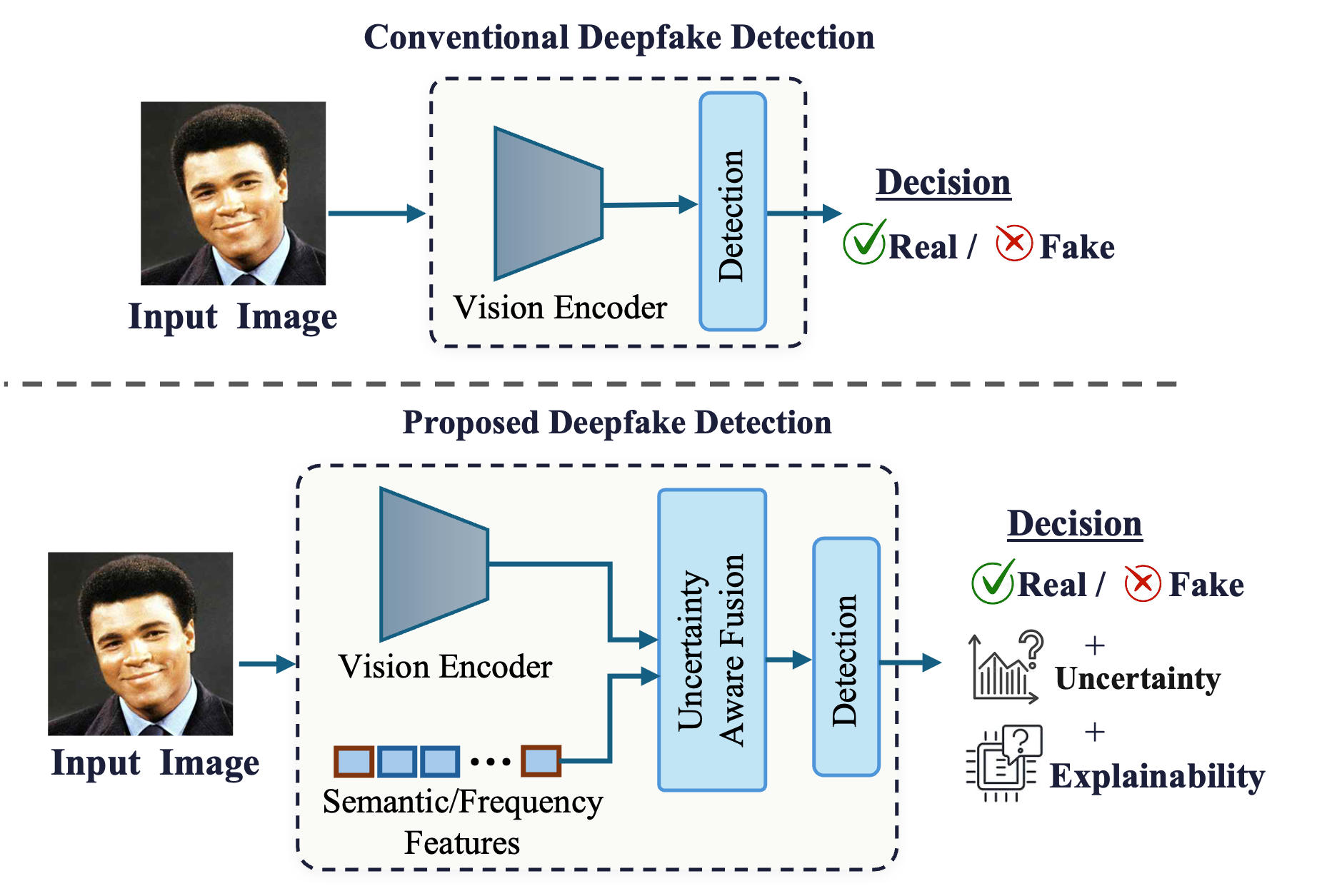}
  \caption{Conventional vs. proposed deepfake detection. Existing methods rely primarily on a single visual encoder and produce a binary prediction.  The proposed framework fuses complementary visual, semantic, and frequency streams under an uncertainty-aware fusion to generate calibrated and explainable predictions.}
  \vspace{-0.5cm}
  \label{fig:architect}
\end{figure}
\section{Introduction}
\label{sec:intro}
Deepfake videos, including face swaps and facial reenactments, have become increasingly realistic due to rapid advances in generative models~\cite{rossler2019faceforensics, dolhansky2020dfdc, li2020celeb}. Recent advances in generative models have made deepfakes increasingly realistic and difficult to distinguish from authentic media~\cite{liu2025review, khan2023battling,zhao2023diffswap,uddin2024counter}. To capture this distinction with improved generalization across diverse manipulation techniques, recent deepfake detectors increasingly leverage large-scale pretrained vision models and foundation representations~\cite{yan2024df40, khan2026psa, tasnim2025ai, farooq2026trace, yermakov2026deepfake, khan2026dual, uddin2026face2parts}. Due to large-scale pretraining, these approaches generally improve cross-dataset generalization compared to artifact-centric detectors. However, despite these advances, a fundamental limitation persists; detectors often exhibit overconfident predictions under distribution shift, unseen manipulation methods, or ambiguous samples~\cite{uddin2023robust, kaur2024deepfake, farooq2025transferable,uddin2025guard}.\\
A fundamental limitation of current detectors is their reliance on a single dominant representation, whether based on appearance artifacts, frequency information, or foundation-model embeddings~\cite{shiohara2022detecting, li2024freqblender, wang2026multi,uddin2025advbench, li2026multiple, xu2026mare, tasnim2026grex, uddin2021analysis}. Consequently, predictions may remain highly confident even when different facial cues provide conflicting evidence, particularly under distribution shift where individual feature spaces become unreliable or capture dataset-specific biases. We argue that reliable deepfake detection requires integrating multiple complementary views of facial content while explicitly modeling disagreement among them as a measure of predictive uncertainty. Authentic faces exhibit consistent relationships across visual appearance, facial behavior, and low-level forensic characteristics, whereas manipulated content often disrupts these natural dependencies. Therefore, modeling these cross-stream inconsistencies provides a more robust signal rather than relying on any single representation alone. \\
From the perspective of trustworthy deepfake detection, the key question is not only whether a model correctly detects manipulated content, but whether its confidence estimates remain reliable under distribution shift and common real-world degradations such as noise, blur, JPEG compression, brightness, contrast, and illumination variations. Existing detectors~\cite{yan2023ucf, yan2024lsda, haliassos2021lips,saeed2026omnidf,uddin2025adversarial, zheng2021ftcn} achieve strong benchmark performance but often become overconfident on unseen manipulations and degraded inputs, limiting their suitability for real-world forensic and biometric applications. We therefore investigate whether agreement across complementary evidence streams improves both robustness and calibration, and whether cross-stream disagreement provides a more reliable uncertainty signal for selective prediction than conventional single-stream detectors~\cite{luo2021generalizing, qian2020thinking, li2024freqblender}.\\
Motivated by this observation, we propose DISCERN (Disagreement-Calibrated Evidential Reasoning Network), an uncertainty-aware framework for deepfake detection that jointly exploits three complementary evidence streams. A visual stream leverages a CLIP-based encoder to capture high-level facial representations, a semantic stream models consistency across facial attributes, including expressions, action units, pose, and geometry through differentiable constraints, and a structural stream learns class-conditioned dependency patterns between semantic and forensic features to identify manipulation-specific inconsistencies.
To effectively integrate these heterogeneous signals, we introduce Inter-Branch Disagreement Calibration (IBDC), a disagreement-aware uncertainty modeling mechanism that directly incorporates inter-stream disagreement into training. IBDC produces higher uncertainty when evidence streams conflict and lower uncertainty when they agree. This design encourages confidence estimates that better reflect the quality and consistency of the underlying evidence, leading to more reliable predictions under distribution shift. The main contributions of this work are as follows:
\begin{itemize}[noitemsep,topsep=2pt]
    \item We propose DISCERN, a multi-stream deepfake detection framework that jointly exploits visual, semantic, and structural evidence to improve robustness under distribution shift.
    \item We introduce a structural dependency modeling strategy that captures class-dependent relationships between semantic and forensic cues, providing complementary information beyond conventional appearance-based representations.
    \item We develop Inter-Branch Disagreement Calibration (IBDC), a disagreement-aware uncertainty modeling mechanism that links predictive uncertainty to conflicts among heterogeneous evidence sources.
    \item We demonstrate through extensive cross-dataset and real-world degradation evaluations that DISCERN improves generalization, robustness, calibration, and selective prediction, highlighting its suitability for trustworthy deepfake detection.
\end{itemize}

%% file: secs/literature_review.tex
\section{Related Work}
\label{sec:related}
\subsection{Conventional Deepfake Detection}\vspace{-0.2cm}
Early detectors focused on low-level artifacts introduced by generative models, including blending inconsistencies, frequency artifacts, and temporal irregularities. While these approaches achieve strong in-domain performance, they degrade significantly under cross-dataset and cross-manipulation settings~\cite{rossler2019faceforensics, khan2024frame, li2020celeb}. \\ 
To improve generalization under distribution shift, prior work explores complementary forensic cues beyond appearance. One line targets low-level residual and frequency-based representations to capture manipulation-agnostic signals~\cite{luo2021generalizing, qian2020thinking,farooq2026trace, uddin2025sheild, li2024freqblender}, while data-centric augmentation strategies expose models to broader manipulation distributions~\cite{li2020face, shiohara2022detecting, farooq2025lightweight, cheng2024prodet,saeed2025realism,farooq2025lightweight}. In parallel, higher-level semantic and temporal modeling incorporates facial motion consistency and expression dynamics~\cite{yan2023ucf, yan2024lsda,farooq2025generalized, khan2023spotnet, haliassos2021lips,khan2022toward, zheng2021ftcn, haq2025domain,raza2024ruleboost}, including AU-based methods for localized inconsistency modeling~\cite{bai2023aunet, jin2025towards}. Despite advances, most methods rely on a single decision pathway that fuses heterogeneous cues into a scalar authenticity score, limiting generalization under conflicting or ambiguous evidence.
 \vspace{-0.2cm}
\subsection{Foundation Models for Deepfake Detection} \vspace{-0.2cm}
Recent work leverages large-scale vision foundation models, particularly CLIP~\cite{radford2021clip,tasnim2026diversity}, to improve cross-domain generalization in deepfake detection. These approaches replace handcrafted features with pretrained embeddings that exhibit stronger cross-model generalizability.\\
To adapt these representations, prior work explores lightweight fine-tuning strategies, such as prompt learning~\cite{khan2024clipping}, adapter-based tuning~\cite{cui2025forensics, lin2024facialcomponent}, token manipulation techniques~\cite{yan2024tokenshuffling}, feature decomposition methods~\cite{yan2024orthogonal,uddin2026transformations, khan2023securing}, and visual reprogramming~\cite{bhattacharya2024reprogramming}. These methods consistently demonstrate improved generalization across datasets, confirming the effectiveness of foundation representations for forensic tasks. However, consistent with the limitation highlighted in the introduction, these approaches still rely on a single embedding space followed by a unified prediction head, where heterogeneous forensic cues are implicitly fused. As a result, while representation quality improves, these models do not explicitly represent disagreement across different evidence types, and their confidence estimates remain vulnerable under distribution shift and unseen manipulations. Recent calibration-based efforts partially address this issue~\cite{jin2025towards, zhang2025choose}, but they typically operate as post-hoc corrections rather than mechanisms integrated into the inference process.
\vspace{-0.2cm}
\subsection{Uncertainty and Multi-View Learning} \vspace{-0.2cm}
A central challenge in modern deepfake detection is overconfident prediction under distribution shift, even when evidence is ambiguous or conflicting, motivating uncertainty-aware learning. Classical approaches include Bayesian neural networks and Monte Carlo dropout for estimating epistemic uncertainty, while Evidential Deep Learning (EDL)~\cite{sensoy2018evidential} models predictions as Dirichlet distributions to provide uncertainty-aware classification without sampling overhead. Multi-view learning extends uncertainty estimation to heterogeneous inputs, where Trusted Multi-View Classification (TMC)~\cite{han2021tmc} and its extensions~\cite{han2022tmcdynamic} fuse evidence using Dempster-Shafer theory, and more recent methods exploit inter-modal disagreement as a reliability signal~\cite{xu2024reliable}. However, these approaches are designed for general multi-modal classification rather than deepfake forensics, and typically aggregate evidence into a final prediction without explicitly modeling inter-representation disagreement as a structural learning signal, limiting their ability to capture subtle forensic inconsistencies and reliable uncertainty.
 \vspace{-0.2cm}
\subsection{Causal Modeling for Generalization} \vspace{-0.2cm} 
To address distribution shift, causal representation learning has been proposed as a principled approach for identifying invariant generative mechanisms~\cite{scholkopf2021causal}. In contrast to correlation-based learning, causal methods aim to capture stable structural dependencies across environments. Recent advances in differentiable structure learning, such as NOTEARS and DAGMA~\cite{bello2022dagma}, enable explicit modeling of dependency graphs in learned representations. While these approaches have influenced robustness research, their application to deepfake detection remains limited, as most forensic models rely on implicitly learned invariance rather than explicit structural modeling of real versus manipulated content. Some recent work incorporates counterfactual reasoning~\cite{yan2023ucf}, but still lacks explicit modeling of dependencies across semantic, visual, and forensic cues. These limitations motivate a unified framework that jointly models cross-representation consistency, structural divergence, and uncertainty under conflicting evidence. In the next section, we introduce a detection framework that addresses these challenges via a multi-stream architecture with disagreement-aware uncertainty modeling.

%% file: secs/proposed_method.tex
\section{Proposed Method} 
\label{sec:method}
This section presents the architectural overview, input representations, evidence streams, uncertainty-aware evidence fusion, and training objective of DISCERN.
\subsection{Architectural Overview} \vspace{-0.2cm}

\begin{figure*}[t]
  \centering
  \includegraphics[width=1.0\linewidth]{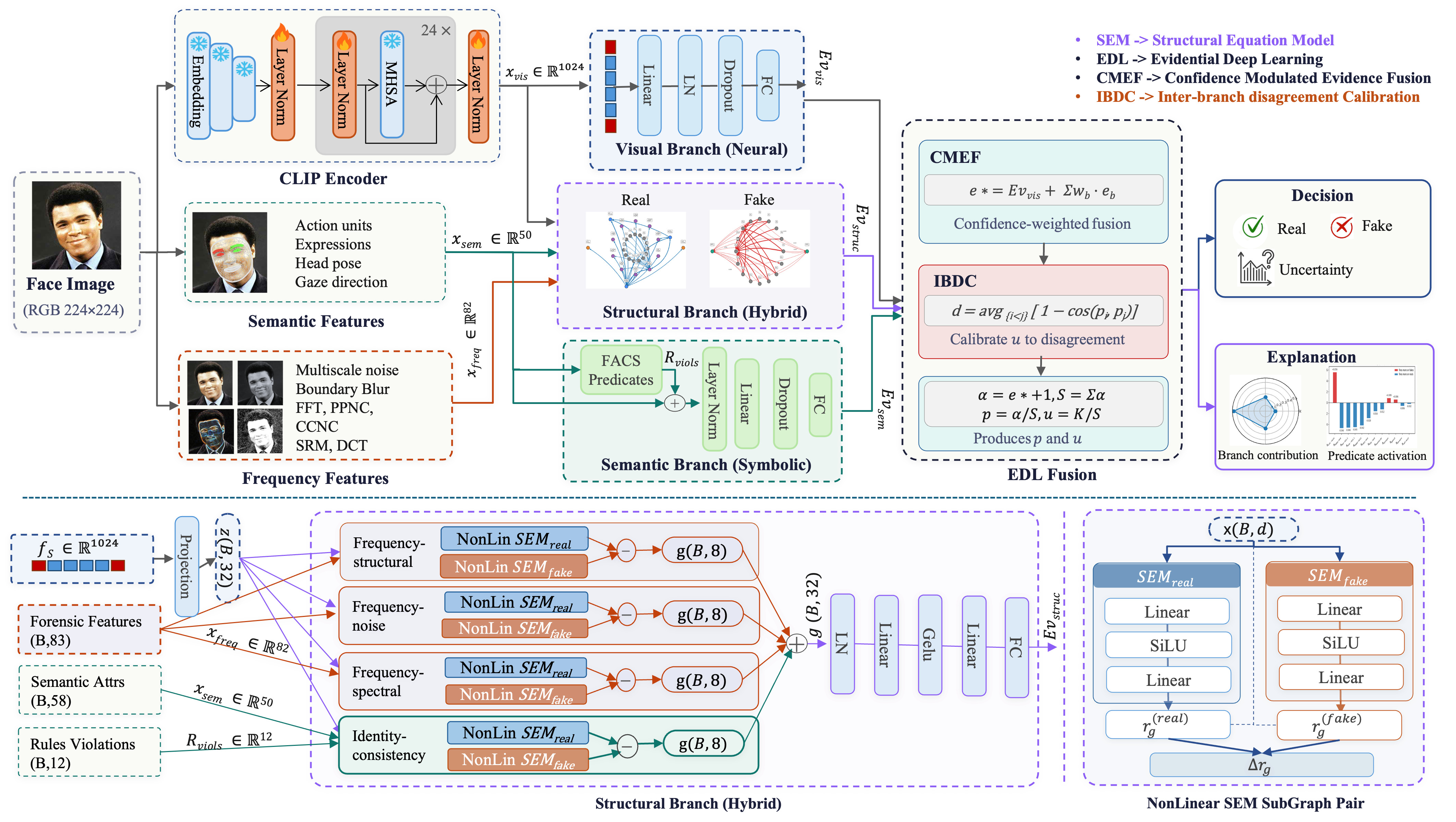}
   \caption{Overview of the proposed framework. A frozen CLIP backbone with layer-norm tuning produces $\bm{Ev}_{\text{vis}}$. A semantic stream evaluates differentiable FACS-grounded predicates over the semantic descriptor $\bm{f}_{\text{sem}}$ to produce $\bm{Ev}_{\text{sem}}$. A structural stream learns four pairs of class-conditioned structural equation models, producing  $\bm{Ev}_{\text{struc}}$. The three streams are fused by EDL into a calibrated decision with epistemic uncertainty $u$.}
  \label{fig:architect}
  \vspace{-0.3cm}
\end{figure*}

We propose an uncertainty-aware deepfake detection framework that identifies manipulations as inconsistencies across complementary representations rather than a single feature space. We hypothesize that deepfakes violate invariant semantic and structural regularities~\cite{yan2023ucf, shiohara2022detecting}, captured through three evidence streams. The visual stream uses a pretrained foundation model~\cite{radford2021clip} for high-level features with strong cross-domain generalization~\cite{cui2025forensics, yermakov2026deepfake}. The semantic stream enforces consistency over facial attributes such as expressions, action units, and geometry~\cite{bai2023aunet, ekman1978facial}, while the structural stream models class-dependent dependencies via structural equation modeling~\cite{bello2022dagma, scholkopf2021causal}.\\
Each stream produces class-wise evidence fused using a Dirichlet-based evidential framework~\cite{sensory2018evidential} to model uncertainty. A disagreement-driven calibration mechanism further links epistemic uncertainty to inter-stream inconsistency~\cite{han2021trusted, xu2024reliable}, ensuring uncertainty reflects conflicting evidence. This yields a unified inference process producing predictive distributions and calibrated uncertainty for out-of-distribution inputs. An overview is shown in Fig.~\ref{fig:architect}.

\begin{table}[t]
\caption{Representative features and predicates used in the proposed approach. Full list in Supplementary 6.A.}
\label{tab:feature_examples}
\centering
\footnotesize
\setlength{\tabcolsep}{4pt}
\begin{tabular}{@{}p{0.28\columnwidth}p{0.62\columnwidth}@{}}
\toprule
\textbf{Feature / Predicate} & \textbf{Description} \\
\midrule
\multicolumn{2}{@{}l}{\textit{Semantic (50)}} \\
\midrule
AU12 intensity & Lip-corner puller activation  in FACS \\
Head yaw & Horizontal head rotation angle \\
\midrule
\multicolumn{2}{@{}l}{\textit{Consistency predicates}} \\
\midrule
$|\text{smile} - \text{AU12}|$ & Expression--AU coherence (smile vs.\ AU12) \\
$|\text{anger} - \text{AU7}|$ & Expression--AU coherence (anger vs.\ AU7) \\
\midrule
\multicolumn{2}{@{}l}{\textit{Frequency structural (24)}} \\
\midrule
Boundary gradient & Sobel magnitude at face boundary \\
Eye-skin color dist. & Chi-squared histogram distance (eye vs.\ skin) \\
\midrule
\multicolumn{2}{@{}l}{\textit{Frequency noise (43)}} \\
\midrule
H. residual STD & SRM horizontal residual standard deviation \\
Cheek noise cons. & Patch-wise noise agreement (cheek regions) \\
\midrule
\multicolumn{2}{@{}l}{\textit{Frequency spectral (15)}} \\
\midrule
HF energy ratio & High-frequency / total FFT energy ratio \\
Eye HF DCT energy & High-frequency DCT energy (eye region) \\
\bottomrule
\end{tabular}
 \vspace{-0.4cm}
\end{table}

\vspace{-0.2cm}
\subsection{Input Representations}  \vspace{-0.2cm}
\label{sec:overview}
Each face crop is represented by three feature groups. The visual representation, $\bm{x}_{\text{vis}} \in \mathbb{R}^{1024}$ is obtained from the class token of a frozen CLIP ViT-L/14 vision encoder~\cite{radford2021clip}. The semantic representation $\bm{x}_{\text{sem}} \in \mathbb{R}^{50}$ encodes facial attributes, including action units, expressions, pose, and geometry~\cite{ekman1978facial, chang2024libreface}. 
All feature extractors are frozen, while the downstream modules are trained.
The semantic features provide structured facial descriptors that remain consistent in authentic images but are often violated by generation pipelines~\cite{bai2023aunet}. Representative examples are shown in Table~\ref{tab:feature_examples}, while the full list is given in Supplementary 6.A.\\
The frequency features $\bm{x}_{\text{freq}} \in \mathbb{R}^{82}$ that include structural, noise, and spectral features capture complementary low-level artifacts across structural, noise-residual, and spectral domains.
Structural features target blending artifacts~\cite{li2020facexray,
shiohara2022detecting}, noise features encode deviations from sensor statistics~\cite{fridrich2012rich,
luo2021generalizing}, and spectral features capture frequency-domain signatures of generative models~\cite{durall2020watch, frank2020leveraging,
qian2020thinking}. Representative examples are shown in Table~\ref{tab:feature_examples}, while the full list is given in Supplementary 6.B.
\subsection{Evidence Streams}
\vspace{-0.2cm}
\label{sec:streams}
The visual stream is based on a visual representation $\bm{x}_{\text{vis}}$ that is mapped through a projection head and linear classifier to produce non-negative evidence, defined as follows:
\begin{equation}
\bm{Ev}_{\text{vis}} =
\operatorname{softplus}\!\bigl(MLP_{\text{vis}}(\bm{x}_{\text{vis}})\bigr).
\end{equation}
where $MLP_{\text{vis}}$ is a projection head followed by a linear classifier, mapping CLIP features to two-class evidence. Softplus ensures non-negativity required by the Dirichlet framework~\cite{sensoy2018evidential}. Following recent foundation-model adaptation strategies ~\cite{yermakov2026deepfake, cui2025forensics}, only $x_{\text{vis}}$ and CLIP's layer-normalization parameters are trained.
The semantic stream is based on $\bm{x}_{\text{sem}}$ and a set of differentiable predicates over
$\bm{x}_{\text{sem}}$ that measure agreement between related signals, including expression-action unit alignment, pose-gaze consistency, and geometric symmetry~\cite{ekman1978facial, bai2023aunet}. Predicate outputs $\bm{\nu} \in [0,1]^{N_r}$ are concatenated with semantic features and mapped to the semantic evidence that is defined as:
\begin{equation}
\bm{Ev}_{\text{sem}} =
\operatorname{softplus}\!\bigl(\mathrm{MLP}_{\text{sem}}([\bm{x}_{\text{sem}} \| \bm{\nu}])\bigr).
\end{equation}
The predicate is selected on the training fold using a validation-based discriminative criterion. From 28 candidates, we retain the top 18. Examples appear in Table~\ref{tab:feature_examples}, while the full set and selection procedure are reported in Supplementary 6.A.
The structural stream models structural dependencies under competing class hypotheses through differentiable structural-equation models~\cite{bello2022dagma, scholkopf2021causal}. For each
sub-domain, we learn two structural-equation models, one conditioned on real and one on fake samples. 

Detection is based on the difference in reconstruction residuals between the two class-conditioned models. For a sub-domain $g$ with input $\bm{u}_g$, we compute sub-graphs as:
\begin{equation}
r_g^{(y)} = \|\bm{u}_g - \mathrm{SEM}_g^{(y)}(\bm{u}_g)\|_2^2,
\end{equation}
and define the residual gap $\Delta r_g = r_g^{(\text{fake})} - r_g^{(\text{real})}$ for each of the
four sub-domains $g \in \{\text{id}, \text{struc}, \text{noise}, \text{spect}\}$. The four gaps are concatenated and mapped to evidence as:
\begin{equation}
\bm{Ev}_{\text{struc}} =
\operatorname{softplus}\!\bigl(\mathrm{MLP}_{\text{struc}}([\Delta r_1 \| \cdots \| \Delta r_4])\bigr).
\end{equation}
This formulation captures shifts in dependency structure rather than individual feature values, an approach motivated by causal representation learning for distribution shift~\cite{scholkopf2021causal}. The learned structures are retained as
interpretable outputs.

\vspace{-0.2cm}
\subsection{Uncertainty-Aware Evidence Fusion}
\vspace{-0.2cm}
\label{sec:nesyedl}
The three streams produce heterogeneous evidence with different inductive biases and signal magnitudes. We fuse them under a Dirichlet-based evidential framework~\cite{sensoy2018evidential} that combines three mechanisms; confidence-modulated evidence fusion, inter-branch
disagreement calibration, and per-stream auxiliary supervision.
Each stream $b \in \{\text{vis}, \text{sem}, \text{struc}\}$ produces a non-negative evidence vector $\bm{Ev}_b$, which defines a per-stream Dirichlet with concentration $\bm{\alpha}_b = \bm{Ev}_b + \bm{1}$ and strength $S_b = \sum_k \alpha_{b,k}$. The visual stream provides the
anchor evidence at full weight, while the semantic and structural streams contribute proportionally to their own strength~\cite{han2021trusted}. For each modulated stream $b \in \{\text{sem}, \text{struc}\}$ we compute a per-sample confidence weight as:
\begin{equation}
\phi_b = \sigma\!\left(\frac{S_b - K}{\tau}\right) \in (0, 1)
\label{eq:phi}
\end{equation}
where $\sigma(\cdot)$ is the sigmoid, $\tau$ a learnable temperature, and $K$ the number of classes. A stream with weak evidence ($S_b \approx K$) receives $\phi_b \approx 0.5$, while a confident stream approaches $\phi_b \to 1$. The fused evidence is defined as:
\begin{equation}
\bm{Ev} = \bm{Ev}_{\text{vis}} +
\!\!\sum_{b \in \{\text{sem}, \text{struc}\}}\!\!
\sigma(\gamma_b)\,\phi_b\,\bm{Ev}_b
\label{eq:cmef}
\end{equation}
with $\gamma_b$ a learnable global gate per stream. This lets the model down-weight a stream when its evidence is weak without removing it from the decision. Applying the same Dirichlet construction to $\bm{Ev}$ yields the predicted probabilities and epistemic uncertainty as:
\begin{equation}
\bm{\alpha} = \bm{Ev} + \bm{1}, \qquad
\hat{p}_k = \frac{\alpha_k}{S}, \qquad
u = \frac{K}{S}
\label{eq:dirichlet}
\end{equation}
where $S = \sum_k \alpha_k$ is the strength of the fused evidence. The uncertainty $u$ falls toward zero as evidence accumulates and approaches one when no stream contributes a signal.

\vspace{-0.2cm}
\subsubsection{Inter-Branch Disagreement Calibration}
\vspace{-0.2cm}
Standard evidential learning ties $u$ to evidence magnitude alone, which can produce overconfident predictions when one stream dominates while others disagree~\cite{xu2024reliable}. To address this, we measure pairwise disagreement between the per-stream Dirichlet means 
$\bm{p}_b = \bm{\alpha}_b / S_b$ over $b \in \mathcal{B} =
\{\text{vis}, \text{sem}, \text{struc}\}$ as:
\begin{equation}
d = \frac{1}{n_{\text{pairs}}}
\sum_{\substack{b, b' \in \mathcal{B}\\ b < b'}}
\bigl(1 - \cos(\bm{p}_b, \bm{p}_{b'})\bigr) \in [0, 1],
\label{eq:disagreement}
\end{equation}
where $n_{\text{pairs}}$ is the number of stream pairs, so that $d$ lies in the unit interval and acts as a probability target. We then calibrate the fused uncertainty $u$ against $d$ with a binary cross-entropy term:
\begin{equation}
\mathcal{L}_{\text{ibdc}} =
-d\,\log u - (1 - d)\,\log(1 - u),
\label{eq:ibdc}
\end{equation}
During optimization, $d$ is detached from the computation graph, such that gradients are propagated only through $u$. Consequently, the loss calibrates the uncertainty estimate to match the observed level of inter-stream disagreement, rather than altering the stream predictions themselves.

\subsection{Training Objective}
\label{sec:training}
The base evidential loss applies to any evidence vector $\bm{Ev}$ and
combines three terms; the Dirichlet type-II maximum-likelihood term
$\mathcal{L}_{\text{nll}}$, an annealed Kullback-Leibler regularizer
$\mathcal{L}_{\text{kl}}$ that pulls the posterior toward the uniform
Dirichlet on incorrect classes~\cite{sensoy2018evidential}, and an
accuracy-versus-uncertainty term $\mathcal{L}_{\text{avu}}$ that penalizes
confident errors and uncertain correct predictions~\cite{krishnan2020improving},
\begin{equation}
\mathcal{L}_{\text{edl}}(\bm{Ev}, \bm{y}) =
\mathcal{L}_{\text{nll}} +
\lambda_{\text{kl}}\,\beta(t)\,\mathcal{L}_{\text{kl}} +
\lambda_{\text{avu}}\,\mathcal{L}_{\text{avu}},
\label{eq:edl}
\end{equation}
where $\lambda_{\text{kl}}$ and $\lambda_{\text{avu}}$ weight the two
regularizers, and $\beta(t) = \min(1, t/T_{\text{anneal}}) \in [0,1]$ is a
linear annealing schedule over training step $t$ that prevents premature
over-confidence early in training.

The visual stream has substantially higher capacity than the semantic and
structural streams and would otherwise dominate the gradient flow. To
keep all streams individually predictive, we apply this evidential loss
independently to each stream's evidence,
\begin{equation}
\mathcal{L}_{\text{aux}} =
\sum_{b \in \mathcal{B}}
\mathcal{L}_{\text{edl}}(\bm{Ev}_b, \bm{y}).
\label{eq:aux}
\end{equation}
This auxiliary supervision prevents the dominant stream from suppressing
the others and ensures that the disagreement signal in
Eq.~\ref{eq:disagreement} reflects genuine stream-level reasoning rather
than degenerate near-uniform outputs.

The overall objective combines the base loss on the fused evidence
(Eq.~\ref{eq:cmef}), per-stream auxiliary supervision (Eq.~\ref{eq:aux}),
disagreement calibration (Eq.~\ref{eq:ibdc}), and the dependency-graph
losses over the four structural sub-domains,
\begin{equation}
\begin{aligned}
\mathcal{L} =
&\,\mathcal{L}_{\text{edl}}(\bm{Ev}, \bm{y}) + \lambda_{\text{aux}}\mathcal{L}_{\text{aux}} + \lambda_{\text{ibdc}}\mathcal{L}_{\text{ibdc}} \\
&+ \lambda_{\text{dag}}\mathcal{L}_{\text{dag}} + \lambda_{\text{div}}\mathcal{L}_{\text{div}} + \lambda_{\text{rec}}\mathcal{L}_{\text{rec}},
\end{aligned}
\label{eq:total}
\end{equation}
where $\mathcal{L}_{\text{dag}}$ aggregates the DAGMA acyclicity
penalty~\cite{bello2022dagma} and $\ell_1$ sparsity over all eight
adjacency matrices (four sub-domains $\times$ two class-conditioned
models), $\mathcal{L}_{\text{div}}$ maximizes the $\ell_1$ distance
between the class-conditioned adjacencies within each sub-domain, and
$\mathcal{L}_{\text{rec}}$ is the label-conditioned reconstruction term on
the structural-equation models.

%% file: secs/experiments.tex
\section{Experiments} \vspace{-0.2cm}
\label{sec:experiments}
\begin{table*}[!t]
  \caption{Calibration (ECE; lower is better) and selective prediction performance (E-AURC $\times 10^{-2}$; lower is better) at the video level for models trained on FF++ (c23). The best and second-best results are highlighted in \textbf{bold} and \underline{underlined}, respectively.}
  \label{tab:calibration}
  \centering
  \footnotesize
  \setlength{\tabcolsep}{6pt}
  \begin{tabular}{@{}l ccccc ccccc@{}}
    \toprule
    & \multicolumn{5}{c}{\textbf{ECE} $\downarrow$} & \multicolumn{5}{c}{\textbf{E-AURC} $\times 10^{-2}$ $\downarrow$} \\
    \cmidrule(lr){2-6} \cmidrule(lr){7-11}
    \textbf{Method}
      & CDFv2 & CDFv3 & DFDC & DFDCP & DFD
      & CDFv2 & CDFv3 & DFDC & DFDCP & DFD \\
    \midrule
    DFD-FCG~\cite{han2025facial}
      & 0.131 & 0.142 & 0.071 & 0.088 & 0.069
      & 4.85 & 4.42 & 8.10 & 6.80 & 6.95 \\
    FSFM~\cite{wang2025fsfm}
      & 0.117 & 0.131 & \underline{0.039} & 0.051 & 0.059
      & 4.21 & 4.17 & \underline{6.37} & 5.07 & 6.30 \\
    Effort~\cite{yan2025effort}
      & \underline{0.029} & 0.085 & 0.060 & \underline{0.041} & \textbf{0.030}
      & \underline{2.11} & 1.98 & 7.78 & 6.27 & \underline{1.20} \\
    GenD~\cite{yermakov2026deepfake}
      & 0.056 & 0.052 & 0.043 & 0.069 & 0.040
      & 2.78 & 1.82 & 7.20 & \underline{4.39} & 1.34 \\
    ForAda~\cite{cui2025forensics}
      & 0.073 & \underline{0.038} & 0.059 & 0.073 & 0.043
      & 3.37 & \underline{1.71} & 7.72 & 6.34 & 2.62 \\
    \midrule
    \textbf{DISCERN (Ours)}
      & \textbf{0.014} & \textbf{0.027} & \textbf{0.035} & \textbf{0.020} & \underline{0.033}
      & \textbf{1.42} & \textbf{1.18} & \textbf{4.68} & \textbf{3.22} & \textbf{1.05} \\
    \bottomrule
  \end{tabular}
  \vspace{-0.4cm}
\end{table*}

We evaluate DISCERN under the cross-dataset generalization protocol
standard in recent deepfake detection literature
\cite{cui2025forensics, yermakov2026deepfake, yan2025effort}.
\vspace{-0.2cm}
\subsection{Training and Testing Configurations}
\vspace{-0.2cm}
\label{sec:setup}
Training is performed on FF++~\cite{rossler2019faceforensics} at
the c23 compression level. Cross-dataset evaluation uses five unseen
benchmarks, namely Celeb-DF-v2 (CDFv2)~\cite{li2020celeb},
Celeb-DF-v3 (CDFv3)~\cite{li2025celebdfpp}, DFDC~\cite{dolhansky2020dfdc}, DFDC-Preview
(DFDCP)~\cite{dolhansky2019dfdcp}, and
DeepFakeDetection (DFD)~\cite{dufour2019dfd}. All datasets use the
DeepfakeBench~\cite{yan2023deepfakebench} preprocessing pipeline of
RetinaFace detection, landmark alignment, $1.3\times$ crop margin,
and $224 \times 224$ resolution. The CLIP ViT-L/14~\cite{radford2021clip} backbone is frozen except for LayerNorm parameters. We train with Adam for 30 epochs at batch size 128 on a single RTX 6000 Ada GPU, with cosine warmup learning rate $3 \times 10^{-4}$ for backbone LayerNorms and $10^{-3}$ for
symbolic and structural streams. Source-paired batches pair each
authentic face with a manipulated counterpart from the same source
video~\cite{yermakov2026deepfake}. Headline numbers are averaged
over three random seeds.\\
We report frame- and video-level AUC, where video scores are obtained by averaging fake probabilities over 32 uniformly sampled frames. Calibration is evaluated using expected calibration error (ECE, 10 equal-width bins) and the confident-wrong rate (CW@0.9), defined as the fraction of incorrect predictions with confidence above 0.9. Selective prediction is measured by excess area Under the risk-coverage curve (E-AURC), which quantifies confidence ranking quality relative to the oracle risk-coverage curve. Confidence is computed as $\max(p_{\text{fake}}, 1-p_{\text{fake}})$ for softmax baselines and using the Dirichlet mean $\mathbf{p}=\boldsymbol{\alpha}/S$ for DISCERN. We further profile inference and adaptation cost (total and trainable parameters, FLOPs, and per-frame latency) against the strongest baselines in Supplementary~6.C, where DISCERN adds negligible overhead over a single CLIP adapter while updating only 3.2\,M parameters.

\vspace{-0.2cm}
\subsection{Calibration and Selective Prediction}\vspace{-0.2cm}
\label{sec:calibration}
Reliable deployment of deepfake detectors requires accurate predictions and well-calibrated confidence estimates. Table~\ref{tab:calibration}
reports ECE and E-AURC for DISCERN against five strong baselines at the
video level. DISCERN attains the lowest ECE on four of the five
benchmarks and the lowest E-AURC on all five. The margins are
substantial: on CDFv2 its ECE of $0.014$ is less than half that of the
best baseline (Effort, $0.029$) and nearly an order of magnitude below
the weakest (DFD-FCG, $0.131$), while its E-AURC is about a third lower
than the next-best ($1.42$ vs.\ $2.11$). Fig.~\ref{fig:reliability}(a)
shows this on CDFv2, where its confidence bins track the diagonal at
every level (ECE $= 0.014$), while GenD exhibits the over-confidence
characteristic of softmax adapters at high confidence, where predictions
concentrate without proportionate accuracy (ECE $= 0.056$).
Fig.~\ref{fig:reliability}(b) orders the detectors by ECE and places
DISCERN first. The only exception is DFD, where Effort achieves a
marginally lower ECE, yet DISCERN retains the lowest E-AURC on that
benchmark, so its confidence ranking remains the more reliable basis for
abstention even when absolute calibration is matched. Under a $10\%$
abstention budget on CDFv2, DISCERN raises AUC from $95.72$ to $97.94$, a
gain of $2.22$ points, against $95.12$ to $95.81$ for GenD softmax
confidence, a gain of $0.69$. This gap indicates that DISCERN
concentrates its errors among low-confidence predictions more effectively
than softmax-margin ranking.
\begin{figure}[!t]
  \centering
    \includegraphics[width=\linewidth]{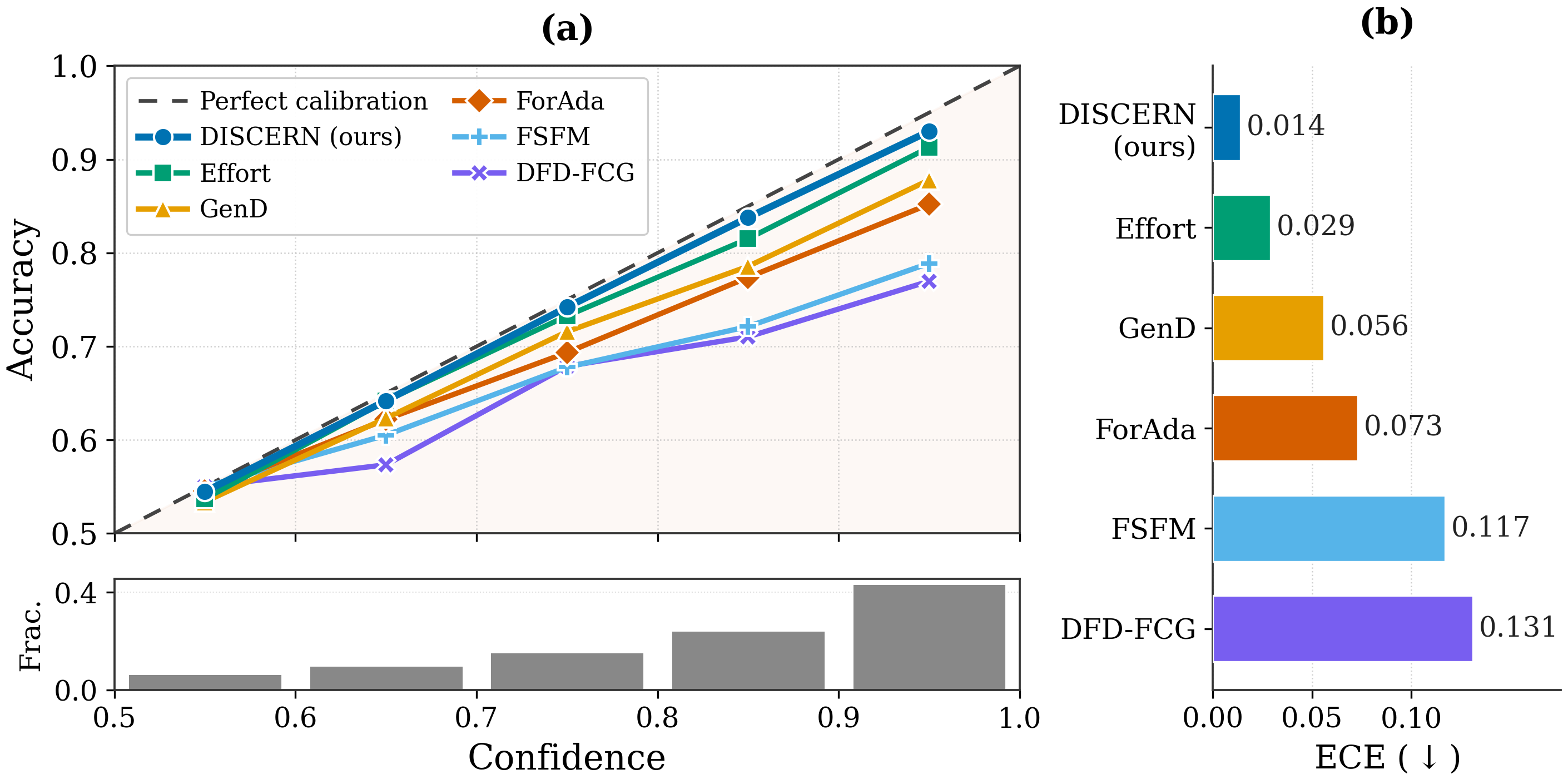}
  \caption{Calibration on CDFv2 (video level). \textbf{(a)} Reliability
diagram of accuracy against confidence, accompanied by the
confidence histogram. The diagonal marks perfect calibration, and
Predictions whose confidence exceeds their accuracy indicate
over-confidence. \textbf{(b)} Per-detector ECE ($\downarrow$). DISCERN
stays closest to the diagonal and attains the lowest ECE.}
\label{fig:reliability}
  \label{fig:reliability}
   \vspace{-0.6cm}
\end{figure}
\begin{table}[b]
  \centering
   \vspace{-0.4cm}
  \caption{Comparison of frame-level AUC (\%) across five datasets under a cross-dataset evaluation protocol, with training performed on FF++ (c23). Best in \textbf{bold} and second-best \underline{underlined}.}
  \label{tab:crossdataset_frame}
  \footnotesize
  \setlength{\tabcolsep}{8pt}
  \begin{tabular}{@{}lccccc@{}}
    \toprule
    \textbf{Method} & CDFv2 & CDFv3 & DFDC & DFDCP & DFD \\
    \midrule
    F3Net~\cite{qian2020thinking}   & 73.5  & --    & 70.2  & 73.5  & 79.8 \\
    SPSL~\cite{liu2021spatial}      & 76.5  & --    & 70.4  & 74.1  & 81.2 \\
    SRM~\cite{luo2021generalizing}    & 75.5  & --    & 70.0  & 74.1  & 81.2 \\
    SBI~\cite{shiohara2022detecting}     & 81.3  & --    & --    & --    & --   \\
    UCF~\cite{yan2023ucf}           & 75.3  & --    & 71.9  & 75.9  & 80.7 \\
    CFM~\cite{luo2023beyond}       & 82.8  & --    & --    & 75.8  & 91.5 \\
    LSDA~\cite{yan2024lsda}      & 83.0  & --    & 73.6  & 81.5  & 88.0 \\
    F2P~\cite{uddin2026face2parts} & 78.63 & 76.35 & 76.98 & 81.52 & 86.73 \\
    FSFM~\cite{wang2025fsfm}      & 85.05 & \underline{84.17} & 80.20 & 85.50 & 88.10 \\
    Effort~\cite{yan2025effort}    & 88.38 & 83.44 & 82.76 & 84.05 & 89.15 \\
    GenD~\cite{yermakov2026deepfake}   & 89.18 & 83.25 & 83.32 & 86.37 & \underline{89.31} \\
    ForAda~\cite{cui2025forensics}   & \underline{89.19} & 81.82 & \textbf{84.29} & \underline{89.01} & \textbf{90.11} \\
    \midrule
    \textbf{DISCERN (Ours)}
     & \textbf{89.33} & \textbf{85.97} & \underline{84.06} & \textbf{91.61} & 88.86 \\
    \bottomrule
  \end{tabular}
   \vspace{-0.6cm}
\end{table}
\subsection{Cross-Dataset Generalization}\vspace{-0.2cm}
\label{sec:crossdataset}
Strong calibration is meaningful only if cross-dataset accuracy is maintained. Tables~\ref{tab:crossdataset_frame} and~\ref{tab:crossdataset_video} compare DISCERN against thirteen state-of-the-art baselines spanning CNN-based~\cite{qian2020thinking, shiohara2022detecting, nguyen2024laanet}, disentanglement~\cite{yan2023ucf, luo2023beyond, yan2024lsda, cheng2024prodet}, CLIP-based~\cite{wang2024videolevel, han2025facial, cui2025forensics, yan2025effort}, and foundation-model methods~\cite{wang2025fsfm, yermakov2026deepfake}, using results reported in the original papers. DISCERN achieves the best frame-level AUC on three of six benchmarks (CDFv2, CDFv3, DFDCP) and the best video-level AUC on three of five (CDFv3, DFDCP, DFD). The largest gains are obtained on Celeb-DF-v3 (+1.80 frame-level, +4.77 video-level) and DFDCP (+2.60 and +1.46), demonstrating improved generalization to diverse manipulation pipelines. Since Celeb-DF-v3 includes 22 generation methods, including diffusion-based synthesis~\cite{li2025celebdfpp}, these results suggest that the structural and semantic streams capture complementary forensic cues missed by single-stream foundation-model adapters. Even on datasets where DISCERN is not the top performer, the margin to the best method remains small (below 0.4 AUC on DFDC at the video level and 1.3 AUC on DFD at the frame level), while consistently ranking among the top two across all benchmarks. These results demonstrate that combining visual, semantic, and structural streams provides more robust cross-dataset generalization than single-stream representations.

\begin{table}[t]
  \centering
  \caption{Comparison of video-level AUC (\%) across five datasets under a cross-dataset evaluation protocol, with training performed on FF++ (c23). Best in \textbf{bold} and second-best \underline{underlined}.}
  \label{tab:crossdataset_video}
  \footnotesize
  \setlength{\tabcolsep}{8pt}
  \begin{tabular}{@{}l ccccc@{}}
    \toprule
    \textbf{Method} & CDFv2 & CDFv3 & DFDC & DFDCP & DFD \\
    \midrule
    F3Net~\cite{qian2020thinking}        & 78.9  & --    & 71.8  & 74.9  & 84.4 \\
    SBI~\cite{shiohara2022detecting}     & 93.2  & --    & 72.4  & 86.2  & 82.7 \\
    CFM~\cite{luo2023beyond}             & 89.7  & --    & --    & 80.2  & --   \\
    LAA-Net~\cite{nguyen2024laanet}      & 95.4  & --    & 86.9  & 86.9  & 98.4 \\
    LSDA~\cite{yan2024lsda}              & 91.1  & --    & 77.0  & --    & --   \\
    ProDet~\cite{cheng2024prodet}        & 92.6  & --    & 70.7  & 82.8  & 90.1 \\
    P\&P~\cite{wang2024videolevel}       & 94.7  & --    & 84.3  & --    & 96.5 \\
    DFD-FCG~\cite{han2025facial}         & 95.0  & --    & 81.8  & --    & --   \\
    F2P~\cite{uddin2026face2parts}  & 85.34 & 83.59 & 84.07 & 87.23 & 89.41 \\
    FSFM~\cite{wang2025fsfm}     & 91.44 & \underline{89.50} & 83.47 & 89.71 & 92.19 \\
    Effort~\cite{yan2025effort}    & 95.6  & 87.88 & 84.3  & 86.17 & 92.76 \\
    GenD~\cite{yermakov2026deepfake}     & \textbf{96.0} & 88.86 & \underline{87.1} & 91.62 & 93.50 \\
    ForAda~\cite{cui2025forensics}       & 95.7  & 87.57 & \textbf{87.2} & \underline{92.89} & \underline{93.95} \\
    \midrule
    \textbf{DISCERN (Ours)}
      & \underline{95.72} & \textbf{94.27} & 86.87 & \textbf{94.35} & \textbf{93.96} \\
    \bottomrule
  \end{tabular}
  \vspace{-0.4cm}
\end{table}
\begin{table}[b]
  \centering
   \vspace{-0.4cm}
  \caption{Video-level AUROC (\%) across FF++ manipulation subsets for cross-manipulation evaluation. Best in
  \textbf{bold}, second-best \underline{underlined}.}
  \label{tab:indomain}
  \footnotesize
  \setlength{\tabcolsep}{5pt}
  \begin{tabular}{@{}lccccc@{}}
    \toprule
    \textbf{Method} & \textbf{DF} & \textbf{F2F} & \textbf{FS} & \textbf{NT} & \textbf{Mean} \\
    \midrule        F2P~\cite{uddin2026face2parts}      & \textbf{99.9}    & 98.4             & \textbf{99.9}    & 94.3             & \underline{98.1} \\
    DFD-FCG~\cite{han2025facial}      & 98.9             & 92.8             & 97.6             & 87.9             & 94.3 \\
    FSFM~\cite{wang2025fsfm}          & 99.2             & 96.8             & 98.9             & 93.4             & 97.1 \\
    ForAda~\cite{cui2025forensics}    & \underline{99.7} & 97.0             & 98.6             & 91.9             & 96.8 \\
    Effort~\cite{yan2025effort}       & 99.4             & 93.2             & 98.4             & 84.6             & 93.9 \\
    GenD~\cite{yermakov2026deepfake}  & 99.5             & \underline{98.1} & 98.7             & \underline{95.5} & 98.0 \\
    \midrule
    \textbf{DISCERN (Ours)}                & 99.4             & \textbf{98.6}    & \underline{99.1} & \textbf{97.2}    & \textbf{98.6} \\
    \bottomrule
  \end{tabular}
  \vspace{-0.4cm}
\end{table}
\subsection{Cross-Manipulation Generalization}
\label{sec:crossmanip}
Beyond cross-dataset transfer, we evaluate in-domain generalization across the four FF++ manipulation types (DF, F2F, FS, NT). Table~\ref{tab:indomain} reports video-level AUROC. DISCERN achieves the best score on F2F and NT and the highest overall mean of 98.6\%, surpassing the second-best mean by 0.5 points. F2P~\cite{uddin2026face2parts} achieves the strongest results on the two face-swap manipulations (DF and FS), where its training protocol specializes the encoder to that family. The largest DISCERN gain is on NT, where it leads the second-best baseline by 1.7 points, and the second-largest is on F2F at 0.5 points. Both are reenactment-style manipulations that disrupt expression-action unit coherence, and the symbolic and structural streams provide complementary evidence in exactly these cases. The consistent performance across all four manipulation types, with no subset below 99.1, confirms that the cross-stream design generalizes across forgery families rather than specializing to a single one.

\begin{figure*}[httb]
  \centering
  \begin{minipage}[t]{0.49\linewidth}
    \centering
    \includegraphics[width=\linewidth]{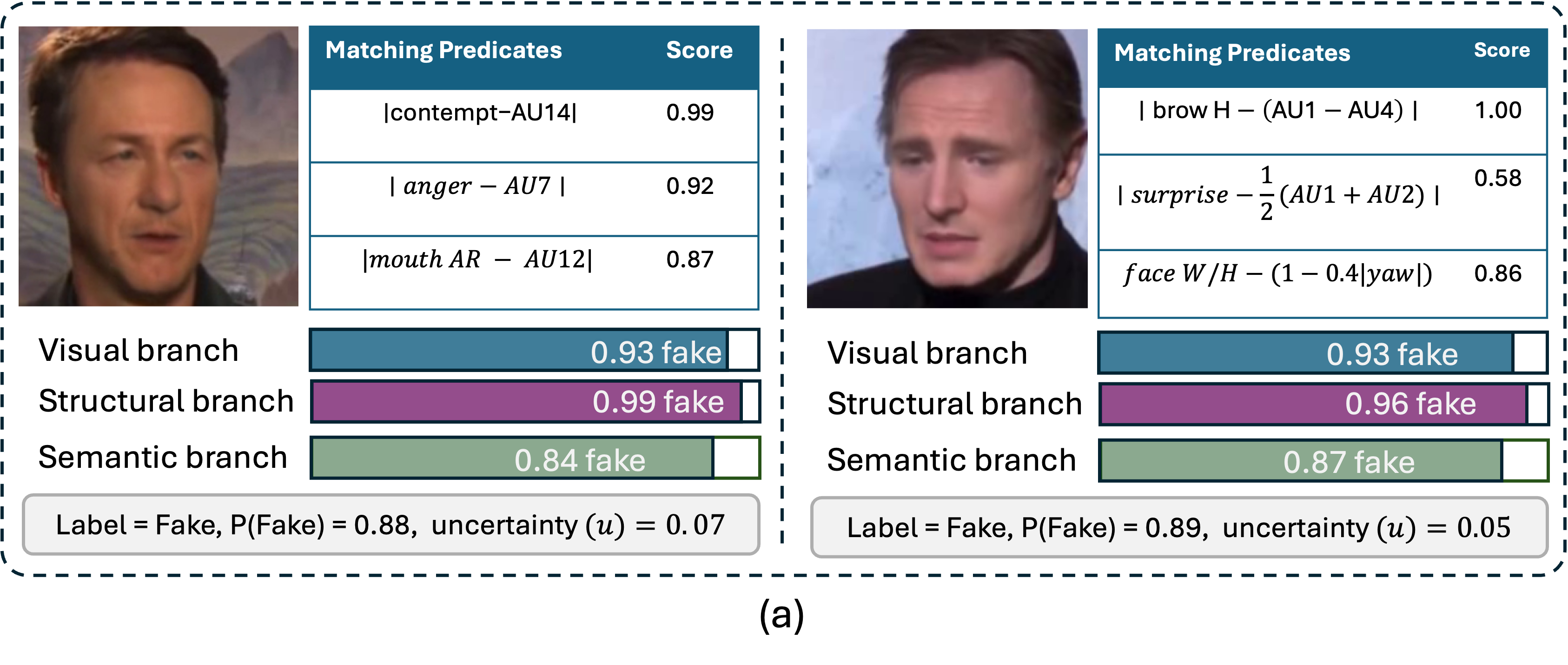}
  \end{minipage}
  \hfill
  \begin{minipage}[t]{0.49\linewidth}
    \centering
    \includegraphics[width=\linewidth]{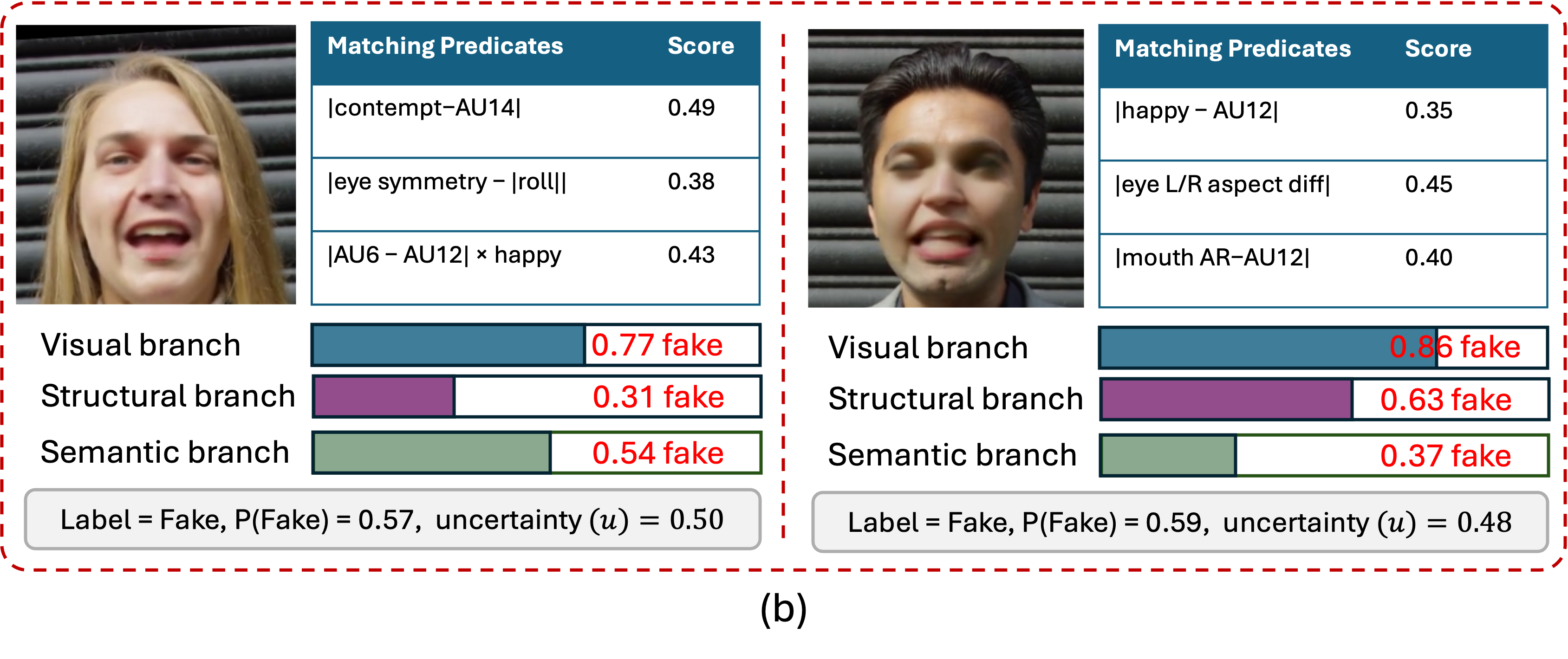}
  \end{minipage}
  \vspace{-0.2cm}
  \caption{Per-prediction outputs on correctly detected fake samples. \textbf{(a)} Confident cases ($u=0.07, 0.05$) with strong, consistent evidence across all three streams. \textbf{(b)} Uncertain cases ($u=0.50, 0.48$) where one stream diverges from the others, and IBDC elevates~$u$.}
  \label{fig:case_studies}
  \vspace{-0.2cm}
\end{figure*}

\subsection{Robustness to Real-World Degradations}
\label{sec:robustness}
To evaluate deployment robustness, we follow the common-corruption protocol of Hendrycks and Dietterich~\cite{hendrycks2019benchmarking} and its deepfake adaptation in DeepfakeBench~\cite{yan2023deepfakebench}. Five corruption types (Gaussian noise, Gaussian blur, JPEG compression, brightness, and contrast) are applied to the CDFv2 test set at five severity levels, and video-level AUC averaged across severities is reported in Table~\ref{tab:corruption_robustness}. All models are trained on FF++ c23 without data augmentation.\\
DISCERN achieves the best AUC on four of the five corruption types and the highest mean AUC of 90.93\%, outperforming DFD-FCG~\cite{han2025facial} by 1.31 points and the strongest CLIP-based baseline ForAda~\cite{cui2025forensics} by 2.67 points. The only exception is Gaussian noise, where DFD-FCG leads by 1.25 points, likely because its graph-based semantic representation is inherently less sensitive to high-frequency additive noise. DISCERN shows the largest gains under JPEG compression (+2.21 over ForAda) and Gaussian blur (+1.97 over DFD-FCG), indicating that its structural and semantic streams remain informative when local visual cues are degraded, while IBDC adaptively emphasizes the most reliable evidence. Performance gains remain consistent under brightness (+1.56 over GenD) and contrast (+1.44 over DFD-FCG) shifts, with DISCERN ranking among the top two across all corruption types, demonstrating robust performance under realistic image degradations.

\begin{table}[t]
  \centering
  \caption{Video-level AUC (\%) on CDFv2 under five common corruptions from the DeepfakeBench~\cite{yan2023deepfakebench} protocol, averaged over five severity levels per corruption. \emph{Mean} is the average across the five corruption columns. Best in \textbf{bold}, second-best \underline{underlined}.}
  \label{tab:corruption_robustness}
  \footnotesize
  \setlength{\tabcolsep}{4pt}
  \begin{tabular}{@{}l ccccc c@{}}
    \toprule
    \textbf{Method}
      & \textbf{Noise} & \textbf{Blur} & \textbf{JPEG}
      & \textbf{Bright.} & \textbf{Contr.}
      & \textbf{Mean} \\
    \midrule
    DFD-FCG~\cite{han2025facial}
      & \textbf{88.43} & \underline{88.46} & 87.32
      & 92.04 & \underline{91.83}
      & \underline{89.62} \\
    FSFM~\cite{wang2025fsfm}
      & 78.34 & 82.61 & 81.95 & 88.17 & 87.41
      & 83.70 \\
    Effort~\cite{yan2025effort}
      & 81.06 & 84.92 & 84.31 & 90.36 & 89.74
      & 86.08 \\
    GenD~\cite{yermakov2026deepfake}
      & 82.47 & 86.13 & 85.62 & \underline{92.18} & 90.47
      & 87.37 \\
    ForAda~\cite{cui2025forensics}
      & 83.62 & 87.05 & \underline{87.84} & 91.62 & 91.18
      & 88.26 \\
    \midrule
    \textbf{DISCERN (Ours)}
      & \underline{87.18} & \textbf{90.43} & \textbf{90.05}
      & \textbf{93.74} & \textbf{93.27}
      & \textbf{90.93} \\
    \bottomrule
  \end{tabular}
  \vspace{-0.1cm}
\end{table}
\vspace{-0.2cm}
\subsection{Ablation Studies}
\vspace{-0.2cm}
\label{sec:ablations}
Table~\ref{tab:ablations} evaluates the contribution of each architectural component on the CDFv2 video-level benchmark. The visual stream with softmax classification achieves an AUC of 91.62 and an ECE of 0.0892, representative of conventional foundation-model adaptation. Replacing softmax with EDL improves calibration by reducing ECE by 28\% with only a 0.78-point drop in AUC, demonstrating the benefit of evidential learning even in a single-stream setting. Integrating the symbolic and structural streams with IBDC further improves performance to 95.72 AUC and 0.0136 ECE. Removing any component degrades both accuracy and calibration. The largest impact comes from removing IBDC, reducing AUC by 2.54 points and increasing ECE by 3.9$\times$, confirming its role in linking inter-stream disagreement to predictive uncertainty. Removing the structural stream decreases AUC by 2.10 points and increases ECE by 3.4$\times$, while removing the symbolic stream reduces AUC by 1.67 points and increases ECE by 2.9$\times$. These results demonstrate that the structural and symbolic streams provide complementary evidence, while IBDC is essential for effectively integrating them into well-calibrated predictions.
\begin{table}[t]
  \centering
  \caption{Component ablation on CDFv2 video level. Each setting
  builds incrementally on the previous one. \cmark{} marks active
  components. All metrics lower-is-better except AUC.}
  \label{tab:ablations}
  \scriptsize
  \setlength{\tabcolsep}{3pt}
  \begin{tabular}{@{}c ccccc cccc@{}}
    \toprule
    \textbf{\#} & \textbf{Vis} & \textbf{EDL} & \textbf{IBDC} & \textbf{Con} & \textbf{Cau}
    & \textbf{AUC} & \textbf{ECE} & \textbf{E-AURC} & \textbf{CW@0.9} \\
    \midrule
    1 & \cmark &        &        &        &        & 91.6  & 0.089  & 4.81 & 0.0212 \\
    2 & \cmark & \cmark &        &        &        & 92.1  & 0.064  & 3.94 & 0.0143 \\
    3 & \cmark & \cmark & \cmark &        &        & 93.0  & 0.051  & 3.21 & 0.0108 \\
    \midrule
    4 & \cmark & \cmark & \cmark & \cmark &        & 93.8  & 0.042  & 2.67 & 0.0078 \\
    5 & \cmark & \cmark & \cmark &        & \cmark & 94.2  & 0.036  & 2.34 & 0.0061 \\
    \midrule
    \textbf{6} & \cmark & \cmark & \cmark & \cmark & \cmark & \textbf{95.72} & \textbf{0.0136} & \textbf{1.42} & \textbf{0.0019} \\
    \bottomrule
  \end{tabular}
  \vspace{-0.4cm}
\end{table}

\vspace{-0.2cm}
\subsection{Analysis}
\vspace{-0.2cm}
\label{sec:explanations}
A central design goal of DISCERN is to provide per-prediction interpretability without post-hoc attribution. Fig.~\ref{fig:case_studies} presents four correctly classified fake samples from FF++ and Celeb-DF-v2, including two low-uncertainty and two high-uncertainty cases. For each prediction, DISCERN reports the most active symbolic predicates with their activation strengths, the evidence from the visual, semantic, and structural streams, and the final uncertainty estimate $u$. The activated predicates correspond to interpretable forensic inconsistencies, such as \emph{``contempt muscle active with neutral expression"} and \emph{``face proportion inconsistent with head yaw"}.\\
Fig.~\ref{fig:case_studies} also illustrates a similar relationship between cross-stream agreement and predictive uncertainty. In Fig.~\ref{fig:case_studies} (a), all three streams consistently support the \emph{fake} class (0.83--0.99), yielding low uncertainty ($u\approx0.06$). In contrast, Fig.~\ref{fig:case_studies} (b) exhibits substantial disagreement: the visual stream strongly favors the \emph{fake} class (0.77 and 0.86), while the semantic or structural stream remains near the decision boundary (0.31 and 0.37), increasing the uncertainty estimate to approximately 0.49 despite the correct prediction. This behavior is consistent with the calibration and selective-prediction results in Table~\ref{tab:calibration}, demonstrating that IBDC effectively links inter-stream disagreement to predictive uncertainty.
\vspace{-0.3cm}
\section{Conclusion}
\vspace{-0.2cm}
In this paper, we presented an uncertainty-aware deepfake detection framework that reformulates detection as identifying inconsistencies across complementary representations rather than relying on a single feature space. By integrating visual, semantic, and structural evidence streams, the proposed framework captures complementary manipulation cues while modeling their disagreement to produce reliable confidence estimates. Central to this framework is Inter-Branch Disagreement Calibration (IBDC), which directly links predictive uncertainty to inter-stream disagreement without requiring post-hoc calibration. Extensive cross-dataset evaluations demonstrate state-of-the-art generalization on unseen benchmarks while consistently improving calibration, achieving strong video-level AUC with low Expected Calibration Error. These results show that combining complementary evidence with disagreement-aware uncertainty provides a robust and trustworthy foundation for deepfake detection under distribution shift, with potential applicability to other high-stakes vision tasks requiring reliable uncertainty estimation.